\documentclass[%
 reprint,
nofootinbib,
 amsmath,amssymb,
 aps,
floatfix,
showkeys
]{revtex4-1}

\usepackage{graphicx}
\usepackage{dcolumn}
\usepackage{bm}

\usepackage{grffile}
\usepackage{float}

\usepackage{amsmath}
 \usepackage{amsthm}
\usepackage{marvosym}
\usepackage{amsfonts}
\usepackage{xcolor}
\usepackage{amssymb}
\usepackage{morefloats}

 \usepackage{wrapfig}
 \usepackage{subfigure}

\usepackage{caption}
\usepackage{hyperref}
\usepackage[ruled,vlined]{algorithm2e}

\usepackage[english]{babel}
\usepackage{xurl}

\newcommand{\twofigl}[5][0]{
\begin{center}
  \vspace{.1cm}
  \includegraphics[width=#3\hsize]{#1}
  \includegraphics[width=#3\hsize]{#2}
   \captionof{figure}{#4}
\label{#5}
  \nobreak
\end{center}}

\newcommand{\diag}{$\textrm{diag}$}

\theoremstyle{plain}

\newtheorem{theorem}{Theorem}[section]

\theoremstyle{definition}
\newtheorem{definition}{Definition}[section]

\newtheorem*{remark}{Remark}

\theoremstyle{definition}
\newtheorem{example}{Example}

\DeclareMathOperator{\dgm}{dgm}

\DeclareMathOperator{\RRips}{\mathbb{R}ips}
\DeclareMathOperator{\Rips}{\mathrm{Rips}}

\DeclareMathOperator{\im}{\mathrm{im}}

\DeclareMathOperator{\CCech}{\check{\mathbb{C}}ech}
\DeclareMathOperator{\Cech}{\check{C}ech}

\DeclareMathOperator{\avg}{\mathrm{Avg}}

\begin{document}




\title{Linguistics from a topological viewpoint}


\author{Rui Dong}
\email{r.dong@hum.leidenuniv.nl}
\affiliation{Centre for Linguistics, Leiden University,\\
Leiden, the Netherlands}



\begin{abstract}
Typological databases in linguistics are usually categorical-valued.
As a result,
it is difficult to have a clear visualization of the data.
In this paper,
we describe a workflow to analyze the topological shapes of South American languages by applying multiple correspondence analysis technique and topological data analysis methods.
\end{abstract}


\maketitle
\tableofcontents



\section{Introduction}
\label{introduction}
Typological data collected by linguists is usually categorical-valued.
For instance,
in the recently released database Grambank \cite{grambank_dataset_zenodo_v1} $189$ out of the $195$ features are binary,
and the rest $6$ features are ternary.
In the Grambank dataset, 
each language is represented by multiple categorical values,
as a result,
it is difficult to measure the difference between two languages.
For instance,
in the case of the commonly used Gower distance,
although it is simple and concise by definition,
the Gower distance does not reflect the frequency of each value in each feature,
which is a crucial characteristic held by a categorical-valued datasets.

In contrast to this,
in this paper we first perform dimensional reduction via the multiple correspondence analysis (MCA) method,
the advantage of the MCA method is that the position of a feature value encodes the information of its frequency inside the dataset,
so that for each language we can form a sub-point cloud consisting of the values of all features.
We then apply the framework of topological data analysis (TDA) to analyze the topological structure of the distributions of each point cloud.
Technically speaking we can use TDA to analyze the topological invariants in any general dimension,
however in practice the meaning of higher dimensional topological invariants is usually vague.
Hence our analysis focuses only on the $1$-dimensional topological invariants in each sub-point cloud,
that is,
the circular structures.

In this paper we restrict ourselves to South American languages.
Specifically, 
we apply the topological methods to analyze the Nuclear-Macro-J\^e (NMJ) family and the Quechuan family. 
In sub-Section \ref{sec:je},
we show that, 
within NMJ, 
there is a significant distinction between languages of the J\^e-proper and the  non-J\^e-proper languages,
and in sub-Section \ref{sec:quech} we show that, within the Quechuan family, there is a significant distinction between northern and southern Quechuan languages.

\section{Related work}
The topological method has already been applied into linguistics,
as is presented in \cite{MR4368966, MR3767893},
where the PCA method is applied to the Syntactic Structures of the World's Languages (SSWL) database 
\footnote{\url{https://ling.yale.edu/syntactic-structures-worlds-language-cross-linguistic-database}}.
In these previous works,
each language is represented as a point inside a Euclidean space,
and the TDA method is applied to analyze the topological structures of different language families.

Our method is inspired by these approaches.
However,
in contrast to these previous works,
in our paper we apply MCA method to display a point cloud for each single language,
so that it is possible to form a visualization for each single language and to compare the shapes of different languages.




\section{Multiple correspondence analysis}
The Multiple correspondence analysis (MCA)\cite{MCA} is a dimension reduction technique that is analogous to principal component analysis (PCA)
for dealing with categorical-valued features instead of quantitative ones.
One main difference between MCA and PCA is that in the case of MCA,
one can obtain either a set of points representing all the samples or a set of points representing all the categorical values.
In our context,
we apply MCA to project all the categorical values into a Euclidean space $\mathbb{R}^d$.
The Appendix \ref{appdx_mca} contains the details of how to implement the MCA in practice.

There are two noteworthy characteristics in MCA:
\begin{itemize}
    \item The distance of a point away from the original point relies on the frequency of the corresponding categorical values. The lower frequency categorical values appear further away from the original point in $\mathbb{R}^d$.
    \item The direction of a point depends on the intrinsic linguistical meaning. In the case of binary features,
    the two categorical values of the same feature lie on a line through the original point of $\mathbb{R}^d$,
    and these two points are in the opposite directions apart from the original point.
\end{itemize}

Hence in the point cloud of all categorical values,
if a point is far away from the original point,
it implies that the representative value is special in the dataset,
the languages that share this specific value should be more relevant.
On the other hand,
if a point is close to the original point,
it means that the corresponding categorical value is commonly appeared in the dataset,
hence in some way we hope to put less weight on it.

This phenomenon suggests that, if  we extract the corresponding values of a specific language out of this point cloud,
we can then obtain a sub-point cloud,
the ``shape'' of which should exhibit some linguistical meaning of this chosen language.
As we have discussed above,
for a chosen language,
if the value of a feature is common,
it will be close to the original point,
and if the value is special,
it will be far away from the original point,
the direction of which relies on the intrinsic meaning from linguistical viewpoint.

More than obtaining only an intuitive visualization of a language,
we would like to be able to investigate the topological structures of the corresponding sub-point clouds.
This is exactly what the framework of topological data analysis is meant to do.

\section{Topological data analysis}
The framework of topological data analysis (TDA) normally starts from a point cloud $P$ inside a Euclidean space $\mathbb{R}^d$.
We can regard the TDA as a higher dimensional generalization of clustering analysis.
That means, 
more than detecting cluster structures of a point cloud, 
the TDA method is able to detect higher dimensional topological structures (e.g., holes, voids, etc.) of the point cloud $P$.
The Appendix \ref{appdx_tda} contains the basic technical definitions and results necessary for this context.
To get a more intuitive understanding of the main idea of TDA,
we start from a simple example of a pretzel-shaped point cloud.
\subsection{A toy example of pretzel}
\begin{wrapfigure}{r}{0.25\textwidth}
    \centering
    \includegraphics[width=0.25\textwidth]{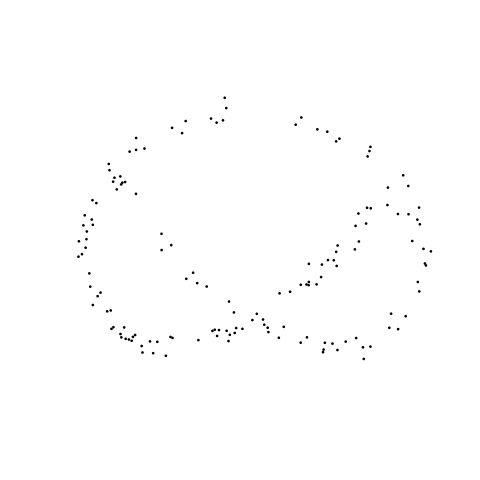}
    \caption{A pretzel-shaped point cloud}\label{img_pretzel_0} 
\end{wrapfigure}

Consider the following pretzel-shaped point cloud in Figure \ref{img_pretzel_0}.
Intuitively,
this point cloud forms three loops.
However, 
since the points are discrete,
the topology of it is trivial,
we need some more technical method to describe the shape of this point cloud.

One way to deal with this issue is to cover each point with a disk of radius $r$, 
as the radius $r$ ranges from $0$ to infinity,
the union of all disks form a connected topological space.

\begin{figure}
    \centering
    \subfigure[]{\includegraphics[width=0.22\textwidth]{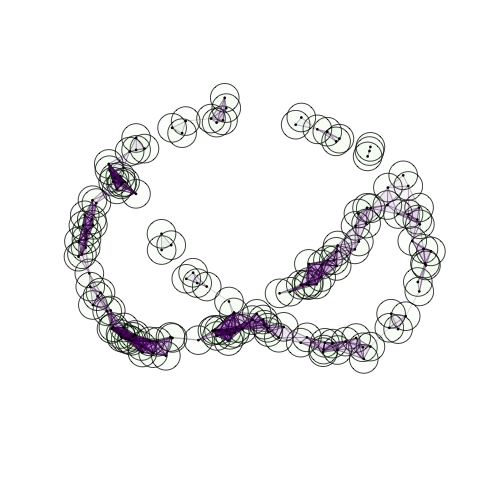}} 
    \subfigure[]{\includegraphics[width=0.22\textwidth]{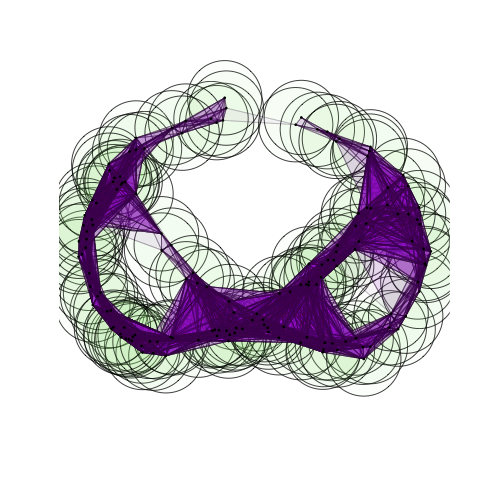}} 
    \subfigure[]{\includegraphics[width=0.22\textwidth]{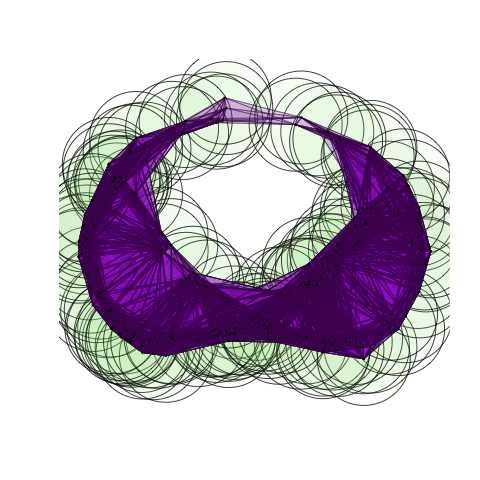}}
    \subfigure[]{\includegraphics[width=0.22\textwidth]{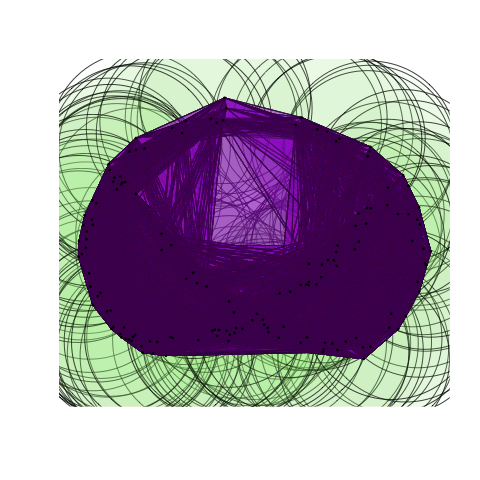}}
    \caption{From (a) to (d): the radius ranges from $0$, the circular structures were born, remained, and dead.}
    \label{img_pretzel_1}
\end{figure}

We display the persistent behaviors of circular structures in Figure \ref{img_pretzel_1},
from which we can observe that as the radius increases,
three circular structures are born (b),
then two of them disappear (c),
finally all disks get intersected together and the third circular structure is dead (d).
This is the core idea of the TDA framework,
as the radius of disks increases,
we can detect the persistent topological structures (such as circular structures) of the point cloud via investigating the union of all the disks.

However,
the union of all the disks (the green objects in Figure \ref{img_pretzel_1}) is not easy to deal with,
to get around this problem,
we construct a ``skeleton'' (the purple objects in Figure \ref{img_pretzel_1}) out of the union of disks.
In terms of TDA, this ``skeleton'' is referred to as ``Vietoris-Rips simplicial complex''\cite{MR2572029}.
One fundamental result in TDA is that the union of all the disks and the ``skeleton'' have the same circular structures (this is referred to as the ``Nerve theorem''\cite{MR2572029} in the framework of TDA.).
Hence,
we can work on the ``skeleton'' instead of the union of all disks.

\subsection{The TDA workflow}

Before describing the framework of TDA,
we first need to recall a topological object called ``$n$-simplex''.
Intuitively speaking,
an $n$-simplex is a generalized ``$n$-dimensional triangle''.
For instance,
a $0$-simplex is a point,
a $1$-simplex is an edge,
a $2$-simplex is simply a triangle,
a $3$-simplex is a tetrahedron.
From these examples,
we observe that one point determines a $0$-simplex,
two points determine a $1$-simplex,
three points determine a $2$-simplex,
four points determine a $3$-simplex,
and in a similar way,
a general $k$-simplex can be defined as an $k$-dimensional object determined by $k+1$ points.
We denote the $k$-simplex determined by the set of points $\{p_0, p_1, ..., p_k\}$ as $\langle p_0, p_1, ..., p_k\rangle$.   

We can then construct multiple simplices out of a point cloud $P$ by the pairwise distances of the points.
In more detail,
we fix a threshold $r>0$,
if the pairwise distances of any $k+1$ points $\{p_0, p_1, ..., p_k\}\in P$ are all less than the threshold $r$,
then these $k + 1$ points determine a $k$-simplex.
The collection of all such simplices determined by the threshold $r$ is referred to as the Vietoris-Rips simplicial complex with respect to the point cloud $P$,
which is denoted as $\Rips_r(P)$.

The object of $\Rips_r(P)$ is still too abstract to deal with,
however,
we can construct a collection of vector spaces out of $\Rips_r(P)$ over the field $\mathbf{F_2}=\{0, 1\}$,
we refer to these vector spaces as spaces of $k$-chains, and denote them as $C_k$ for $k \geq 0$.
In more detail,
a $k$-chain $\gamma$ is a formal finite sum that looks like:
$\gamma=c_1\sigma_1 + c_2\sigma_2 +...+ c_n\sigma_n$,
with $c_i = 0 \textrm{ or }1$ and $\sigma_i$ being a $k$-simplex,
and the vector space $C_k$ contains all such objects.

\begin{wrapfigure}{r}{0.23\textwidth}
    \centering
    \includegraphics[width=0.19\textwidth]{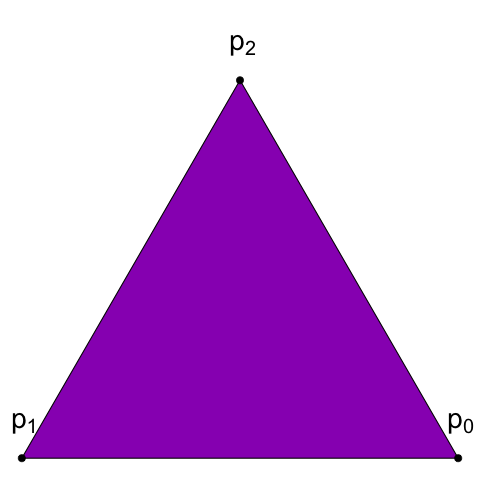}
    \caption{A $2$-simplex $\langle p_0, p_1, p_2\rangle$}\label{trig} 
\end{wrapfigure}

It is obvious that the simplices are not independent to each other,
for example,
if $\Rips_r(P)$ contains a $2$-simplex $\langle p_0, p_1, p_2\rangle$ as is displayed in Figure \ref{trig},
it indicates that all the pairwise distances $d(p_0, p_1)$,  $d(p_1, p_2)$ and $d(p_2, p_0)$ are less than or equal to the threshold $r$,
hence $\Rips_r(P)$ definitely contains the $1$-simplices $\langle p_0, p_1\rangle$, 
$\langle p_1, p_2 \rangle$ and $\langle p_2, p_0\rangle$,
and the $0$-simplices $\langle p_0\rangle$, $\langle p_1\rangle$, $\langle p_2\rangle$ as well.

The relationship between $k$-simplices and 
$(k-1)$-simplices is described by the so-called boundary operator $\partial_{k}$ which extracts the boundaries of all $k$-simplices.
In the case of $2$-simplex $\langle p_0, p_1, p_2\rangle$ as displayed in Figure \ref{trig},
the boundary contains three $1$-simplices $\langle p_0, p_1\rangle$, $\langle p_1, p_2 \rangle$ and $\langle p_2, p_0\rangle$,
we can describe it
with the following equation:
\begin{equation}\label{eqn_bd_1}
\partial_2(\langle p_0, p_1, p_2\rangle)
=
\langle p_0, p_1\rangle + \langle p_1, p_2 \rangle + \langle p_2, p_0\rangle.
\end{equation}

One significant property of the boundary operators is that if you take the boundary of a simplex twice,
you always get $0$.
In other words,
\begin{equation}\label{eqn_bd_sqr}
\partial_{k-1}\circ \partial_{k}=0.
\end{equation}

To explain it,
we still take the $2$-simplex $\langle p_0, p_1, p_2\rangle$ as an example.
We claim that $\partial_1\circ \partial_2(\langle p_0, p_1, p_2\rangle)=0$. 
In fact,
for a $1$-simplex $\langle p_i, p_j\rangle$,
\begin{equation}\label{eqn_bd_2}
\partial_1\langle p_i, p_j\rangle
=
\langle p_i\rangle + \langle p_j\rangle,
\end{equation}
combining Eqn \eqref{eqn_bd_1} and Eqn \eqref{eqn_bd_2} together,
we obtain that 
$$
\partial_1\circ \partial_2(\langle p_0, p_1, p_2\rangle)
=
2 \langle p_0\rangle + 2 \langle p_1\rangle + 2 \langle p_2\rangle.
$$
Remind that in our computation of boundary operators,
all the coefficients are over the field $\mathbf{F}_2$,
that is,
$2=0$,
therefore,
$\partial_1\circ \partial_2(\langle p_0, p_1, p_2\rangle)=0$. 

According to the Eqn \eqref{eqn_bd_sqr},
all the images of the boundary operator $\partial_{k-1}$ lie inside the kernel of the boundary operator $\partial_k$,
i.e.,
$\im(\partial_{k-1})\subset \ker(\partial_k)$.
We then define the Betti number $\beta_k$ as the difference of dimensions of $\ker(\partial_k)$ and $\im(\partial_{k-1})$,
that is,
$\beta_k = \dim(\ker(\partial_k)) - \dim(\im(\partial_{k-1}))$.
Intuitively speaking,
the Betti number $\beta_k$ gives the counting of $k$-dimensional structures.
More precisely,
$\beta_0$ counts the connected components, 
$\beta_1$ counts the circular structures,
$\beta_2$ counts the void structures, 
and $\beta_k$ counts the $n$-dimensional topological structures.

\begin{remark}
In the context of algebraic topology,
the quotient space $\ker(\partial_k) / \im(\partial_{k-1})$ is called the $k$-th homology,
and $\beta_k$ is just the dimension of the $k$-th homology.
\end{remark}

In our context,
we focus mainly on the first Betti numbers $\beta_1$,
i.e.,
the counting of circular structures.
To explain $\beta_1$ more explicitly,
we show how to compute $\beta_1$ of  the boundary of the $2$-simplex in the following example.
\begin{wrapfigure}{r}{0.23\textwidth}
    \centering
    \includegraphics[width=0.19\textwidth]{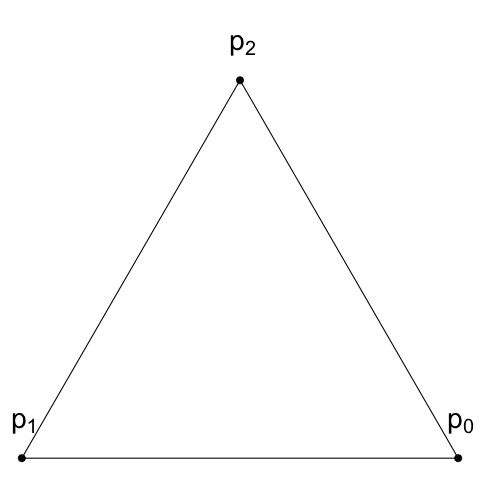}
    \caption{The boundary of the $2$-simplex $\langle p_0, p_1, p_2\rangle$.}\label{trig_bd} 
\end{wrapfigure}
\begin{example}
Consider the boundary of the $2$-simplex $\langle p_0, p_1, p_2\rangle$ as is displayed in Figure \ref{trig_bd}.
According to the definition of simplex,
it contains three $1$-simplices: $\langle p_0, p_1\rangle$, $\langle p_1, p_2\rangle$, and $\langle p_2, p_0\rangle$,
thus the space of $1$-chains $C_1$ is a $3$-dimensional vector space over $\mathbf{F}_2$.
We then need to find the kernel of $\partial_1$,
that is,
to find out all the solutions of the equation:
$$\partial_1(c_1\langle p_0, p_1\rangle + c_2\langle p_1, p_2\rangle + c_3\langle p_2, p_0\rangle)=0.$$
It is easy to see that the only solution is $c_1=c_2=c_3=1$,
i.e.,
$\ker(\partial_1)=\{\langle p_0, p_1\rangle + \langle p_1, p_2\rangle + \langle p_2, p_0\rangle\}$ and therefore $\dim(\ker(\partial_1))=1$.
We then need to find out $\dim(\im(\partial_2))$,
however,
there is no $2$-simplex in the boundary of $\langle p_0, p_1, p_2\rangle$,
hence the space of $2$-chains $C_2$ is a zero dimensional vector space,
and hence $\dim(\im(\partial_2))=0$.
Therefore the Betti number $\beta_1=1-0=1$ and we show that the boundary of a $2$-simplex $\langle p_0, p_1, p_2\rangle$ has one circular structure.
\end{example}
As the threshold $r$ increases from $0$,
we can obtain a series of different Vietoris-Rips simplicial complex $\{\Rips_r(P)\}_{r\geq 0}$.
Roughly speaking,
the procedure of TDA is to study the persistent behavior of Betti numbers with respect to the threshold $r$.

The family of Betti numbers $\beta_k$ parameterized by the threshold $r$ is referred to as the $k$-th persistent Betti numbers.

\begin{remark}
In terms of TDA, 
the $k$-th persistent Betti numbers are the dimensions of the $k$-th persistent homology parameterized by the threshold $r$.
\end{remark}
More than just counting topological structures with persistent Betti numbers,
we can detect at which threshold values a topological structure is born and dead.
\begin{wrapfigure}{r}{0.24\textwidth}
    \centering
    \includegraphics[width=0.2\textwidth]{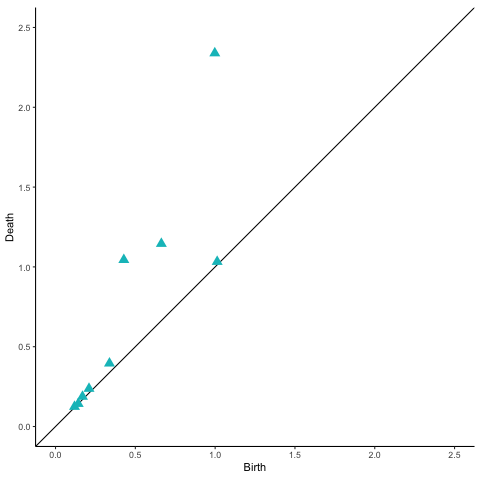}
    \caption{The dim-$1$ persistence diagram of the pretzel-shaped point cloud}\label{img_pretzel_dg} 
\end{wrapfigure}
If for each topological structure we record the threshold values at which it is born and dead,
and plot them in a two-dimensional diagram,
we can then have a visualization of the topological structures of the point cloud $P$.
More precisely,
for instance,
if when the threshold $r$ is equal to $b$,
a topological structure is born,
and when $r$ is equal to $d$, this topological structure is dead,
we then record the point $(b, d)$ on the diagram.
We refer to this diagram as a persistence diagram.
In the case of pretzel-shaped point cloud (Figure \ref{img_pretzel_0}),
the persistent diagram with respect to the first persistent Betti numbers is displayed in Figure \ref{img_pretzel_dg}.
We can see that there are three remarkable points that are far away from the diagonal line,
which represent the three circular structures in Figure \ref{img_pretzel_0}.
The rest of the points in Figure \ref{img_pretzel_dg} are closer to the diagonal line,
which means their surviving time is much shorter,
and they are treated as noise.

\begin{remark}
In the context of MCA,
the more common feature values will be concentrated around the original point,
there would be less chance for them to form some significant circular structures
(In an extreme case,
if all the samples share the same value for each feature,
all the feature values in the MCA plot will collapse into the original point,
hence no any circular structure will be formed.).
Even though any circular structure appears,
it would be gone in a very short time,
which should be treated as noise.
On the other hand,
since those low frequency feature values are further away from the original point,
if any circular structures could be constructed from them, 
such circular structures will be survived for longer time and they should be more meaningful in the framework of TDA.
\end{remark}

We hope to quantify the difference between two point clouds by comparing their corresponding persistence diagrams,
however,
it is usually difficult to compare two persistence diagrams by visualization,
therefore we also need some suitable methods to measure the difference between different persistence diagrams,
by ``suitable'' we mean a ``slight'' fluctuation of sub-point clouds should result in only a ``slight'' difference between the corresponding persistence diagrams 
(Technically, 
this is referred to as ``Stability Theorem'' in TDA
).
Fortunately there are many such suitable options,
one option is the $p$-Wasserstein distance with $p>0$ being a parameter.
Especially when $p=\infty$,
we call the $\infty$-Wasserstein distance the bottleneck distance.
We skip the exact definition of $p$-Wasserstein distance here since it is too technical,
the readers can check Definition \ref{dfn_wass} in Appendix \ref{appdx_tda} for details.
Once we compute out the (topological) distances between each pair of languages,
many different data analysis methods such as clustering analysis, kernel methods, multidimensional scaling (MDS) can then be applied.
In sub-Section \ref{sec:je} and sub-Section \ref{sec:quech} we will apply the MDS to display a visualization of two South American language families: Nuclear-Macro-J\^e and Quechuan.

\section{Data analysis procedure}
In this paper,
we analyze\footnote{A repository of the images used for this paper is available at $\href{https://github.com/ruidongsmile/Lingusitics_vs_topology}{https://github.com/ruidongsmile/Lingusitics\_vs\_topology}$} only the languages of South America in  Grambank(v1.0.3) \cite{grambank_dataset_zenodo_v1}.
The Grambank dataset is the largest comparative grammatical database available. It covers $2467$ language varieties and a wide range of grammatical phenomena in $195$ core features,
thus providing morphosyntactic profiles of languages across several linguistic subdomains (it lacks information on phonology) \cite{grambank}. The data, coded by a large team of linguists, are freely accessible at \url{https://grambank.clld.org/}.
Almost all the features in Grambank are binary, 
the only six ternary features (GB024, GB025, GB065, GB130, GB193, GB203) all concern word order:
``what is the order of elements X and Y?'' with the alternatives ``XY'', ``YX'', 
 or ``both''.
There are many different ways to analyze discrete data, 
the MCA method is one suitable option for us.

\subsection{Data preprocessing}
We first clean the dataset using R package ``glottospace''\cite{glottospace}.
The Grambank dataset contains $224$ languages of South America covering $41$ different language families and $30$ isolates,
among which $217$ languages are classified as ``language'' and $7$ languages are classified as ``dialect''. 
To keep the analysis unbiased, 
we keep only one dialect per language.
As a result, 
the languages 
katu1276
and
tere1281
are removed.
Since we restrict our analysis to South American languages,
some features (GB165,GB291,GB319,GB320) contain only one unique value,
which do not play a significant role and are ignored in our analysis.

Another issue of the Grambank database is that the values of many features are not known.
For instance,
in the case of language 
Suru\'i (suru1262),
$180$ out of $195$ features are missing.  
To make imputation not excessive,
we remove features and languages with more than $20\%$ missing values.
We then apply random forest algorithm to impute the rest missing data implemented by the R package ``mice'' \cite{mice, VANBUUREN2018}.
Finally we split each ternary feature into two binary features,
as the form of ``Is the order XY?'' and ``Is the order YX?''.
After the data preprocessing procedure,
we obtain $183$ languages and $60$ binary features.

\subsection{MCA processing}
\begin{figure}
    \centering
    \includegraphics[width=0.4\textwidth]{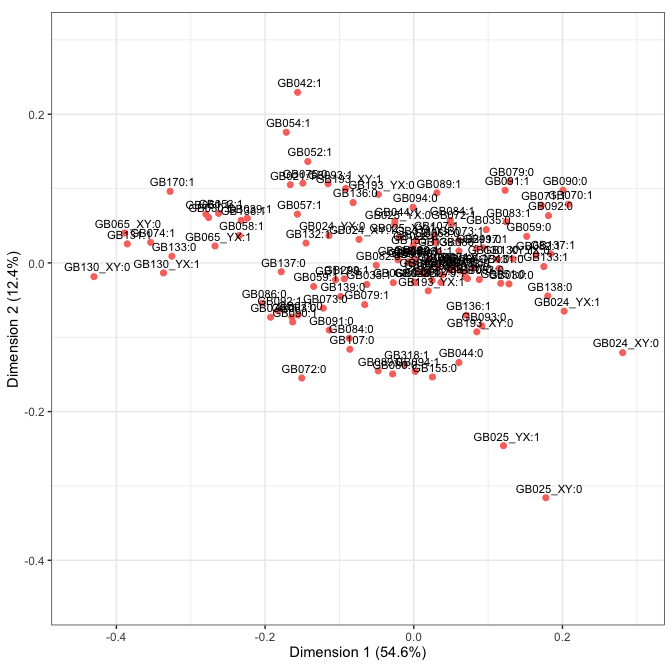}
    \caption{The first two components of MCA projection. Each point represents a feature value.}\label{mjca_ggplot}
\end{figure}

We implement the MCA algorithm using the R package ``ca'' \cite{ca_pkg},
here we choose the method ``adjusted'' \cite{mjca} so as to get a better approximation.
\begin{wrapfigure}{r}{0.22\textwidth}
    \centering
\includegraphics[width=0.21\textwidth]{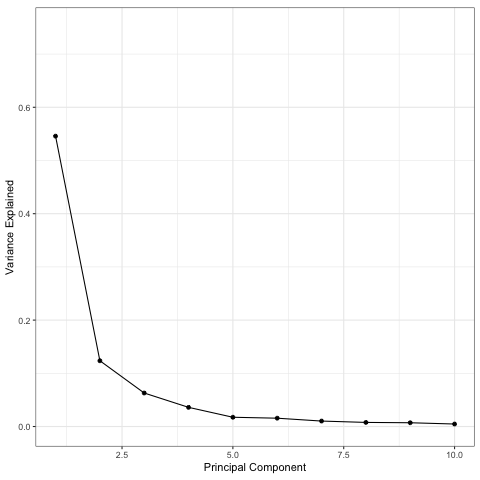}
    \caption{The Scree plot of variance.}\label{fig:scree_plot}
\end{wrapfigure}
The scree plot of variance is displayed in Figure \ref{fig:scree_plot}.
The first two components account for $66.9\%$ of the total variance,
we plot the corresponding two-dimensional point cloud in Figure \ref{mjca_ggplot}.
Here each point represents a feature value.
For instance,
the point ``GB054:1'' represents the value $1$ of feature GB054.
Since in our context we have $60$ binary features,
hence the point cloud obtained via MCA contains $120$ points,
and each sub-point cloud contains $60$ points corresponding to the values of the $60$ features.

\subsection{The shapes of languages}
\begin{figure}
    \centering
    \includegraphics[width=0.4\textwidth]{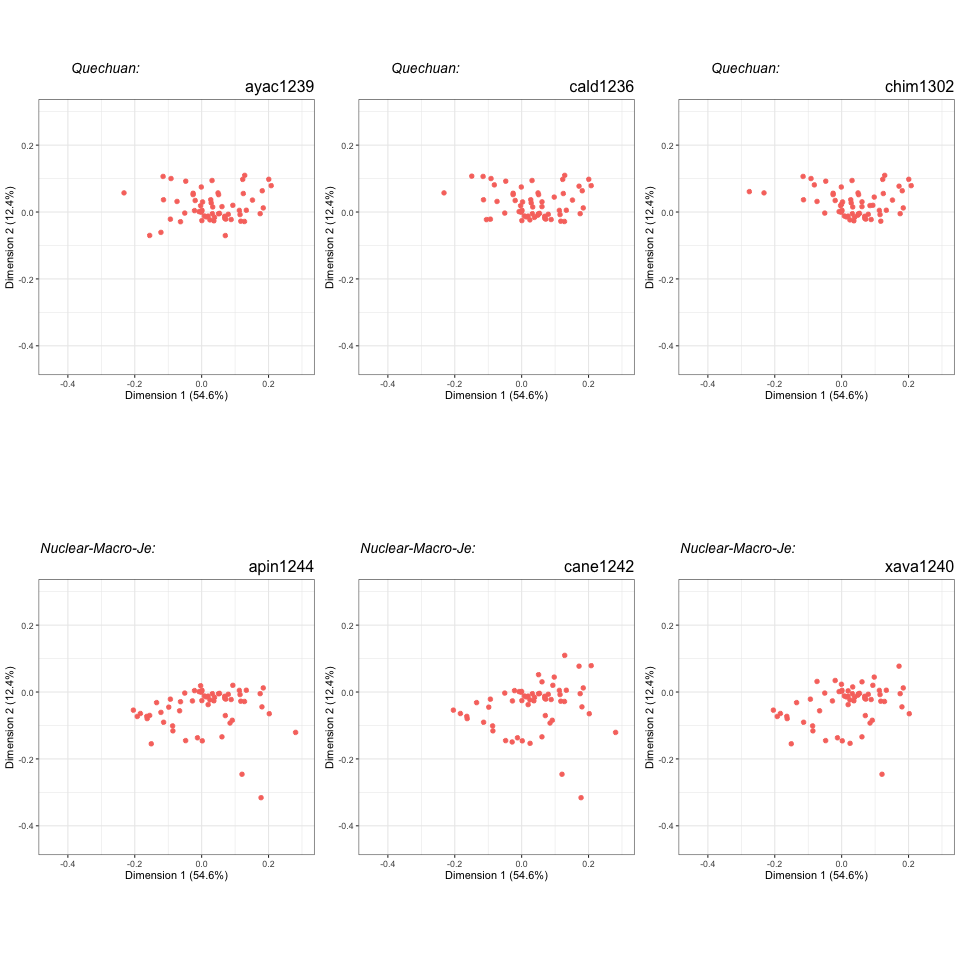}
    \caption{The sub-point clouds of some languages from families Quechuan and Nuclear-Macro-J\^e.}
    \label{fig:family_clouds}
\end{figure}

\begin{figure}
    \centering
    \includegraphics[width=0.4\textwidth]{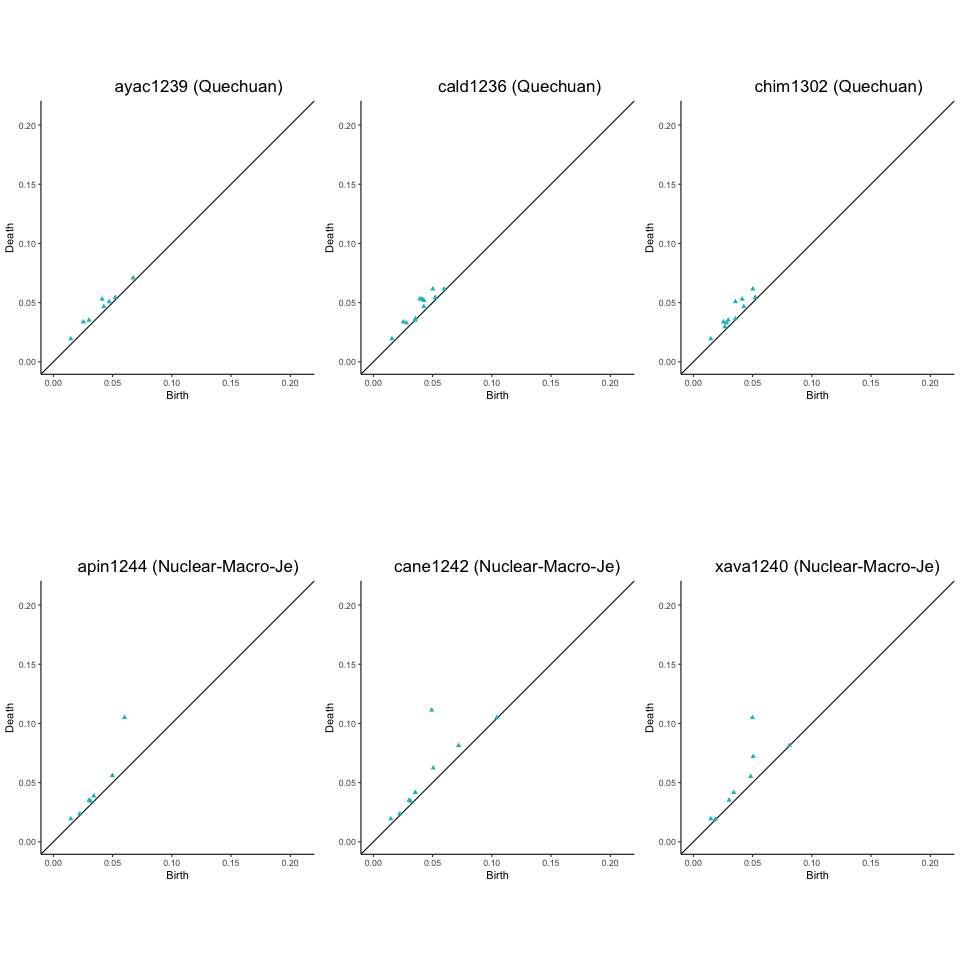}
    \caption{The persistence diagrams with respect to circular structures of some languages from families Quechuan and Nuclear-Macro-J\^e.}\label{fig:family_pd}
\end{figure}

We can now visualize each language by filtering out the feature values.
We pick up three languages from each family of Quechuan and Nuclear-Macro-J\^e (NMJ) as examples,
their corresponding sub-point clouds are displayed in Figure \ref{fig:family_clouds}.
We can see that the languages in the same family have similar shapes.
Specifically,
the sub-point clouds of languages in NMJ family have a circular structure,
while there is not such a circular structure in the languages of the Quechuan family.
The persistence diagrams of corresponding sub-point clouds are displayed in Figure \ref{fig:family_pd}.
We observe that in the persistence diagrams of languages from Quechuan family (the first row in Figure \ref{fig:family_pd}), 
there is not any point significantly away from the diagonal line, 
while in the case of NMJ family (the second row in Figure \ref{fig:family_pd}),
there is a remarkable point away from the diagonal line,
which means that there is a significant circular structure in the corresponding sub-point cloud.
This phenomenon implies that the shape of a sub-point cloud encodes some characteristic of the corresponding language, 
although further research is needed to understand the linguistic meaning.

\section{Applications of TDA}
In this section we show how to apply the TDA method to analyze the Nuclear-Macro-J\^e (NMJ) family and the Quechuan family.
According to the scree plot (Figure \ref{fig:scree_plot}) obtained from the MCA method,
we pick the first four components, 
which account for $76.8\%$ of the total variance,
and we analyze the circular structures (the first homology in the term of topology) of each sub-point cloud.
In this case,
all the sub-point clouds will be four-dimensional,
and it is impossible to visualize them directly,
however we can still detect there circular structures via TDA.

\subsection{Nuclear-Macro-J\^e}\label{sec:je}

Despite some relatively recent advances, Nuclear-Macro-J\^e (NMJ) remains one of the lesser understood families of South America. The family is generally represented as having a rake-like structure, consisting of a number of first-order members, some of which have further branches. The first-order branch for which most reconstructive work has been done is the J\^e branch (or J\^e-proper) \cite[e.g.][]{nikulin_proto_mj}, for which there is general consensus that they form a genealogical sub-unit, in turn divided into a northern, central, and southern branch. For the other first-order branches, the support ranges widely \cite{michael_classifications}, but the languages in our sample that are not from J\^e-proper are generally accepted as part of NMJ. 

We will apply the TDA method to analyze the difference between J\^e-proper languages and the other NMJ languages.
After the data-cleaning processing,
there are $11$ NMJ languages remained (Grambank contains $13$ languages of NMJ in total),
$5$ of which belong to the J\^e-proper branch. The sample languages are divided as follows over the first-order branches (classification follows \cite{glottolog}):

\begin{itemize}
    \item \textsc{J\^e-proper}: Apinay\'e (apin1244), Kanela-Krah\^o (cane1242), Panar\'a (pana1307), Xav\'ante (xava1240), Xokleng (xokl1240).
    \item \textsc{Jabut\'i}: Djeoromitxi (djeo1235)
    \item \textsc{Karaj\'a}: Karaj\'a
    \item \textsc{Maxakali-Borum}: Borum/Kren\'ak (kren1239), Maxakal\'i (maxa1247)
    \item \textsc{Ofay\'e}: Ofay\'e (ofay1240)
    \item \textsc{Rikbaktsa}: Rikbaktsa (rikb1245)
\end{itemize}

\begin{figure}
    \centering
    \includegraphics[width=0.48\textwidth]{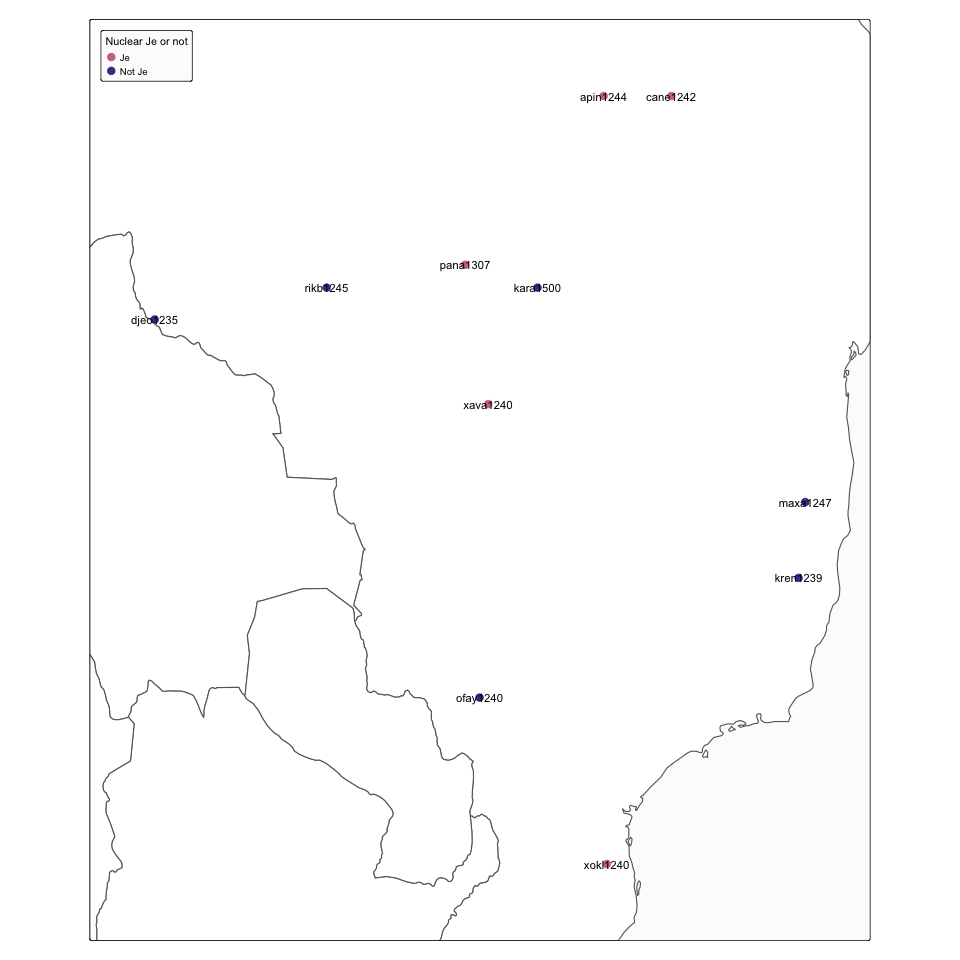}
    \caption{Geographical map of Nuclear-Macro-J\^e (NMJ)}
    \label{fig:map_is_je}
\end{figure}

\begin{figure}
    \centering
    \includegraphics[width=0.4\textwidth]{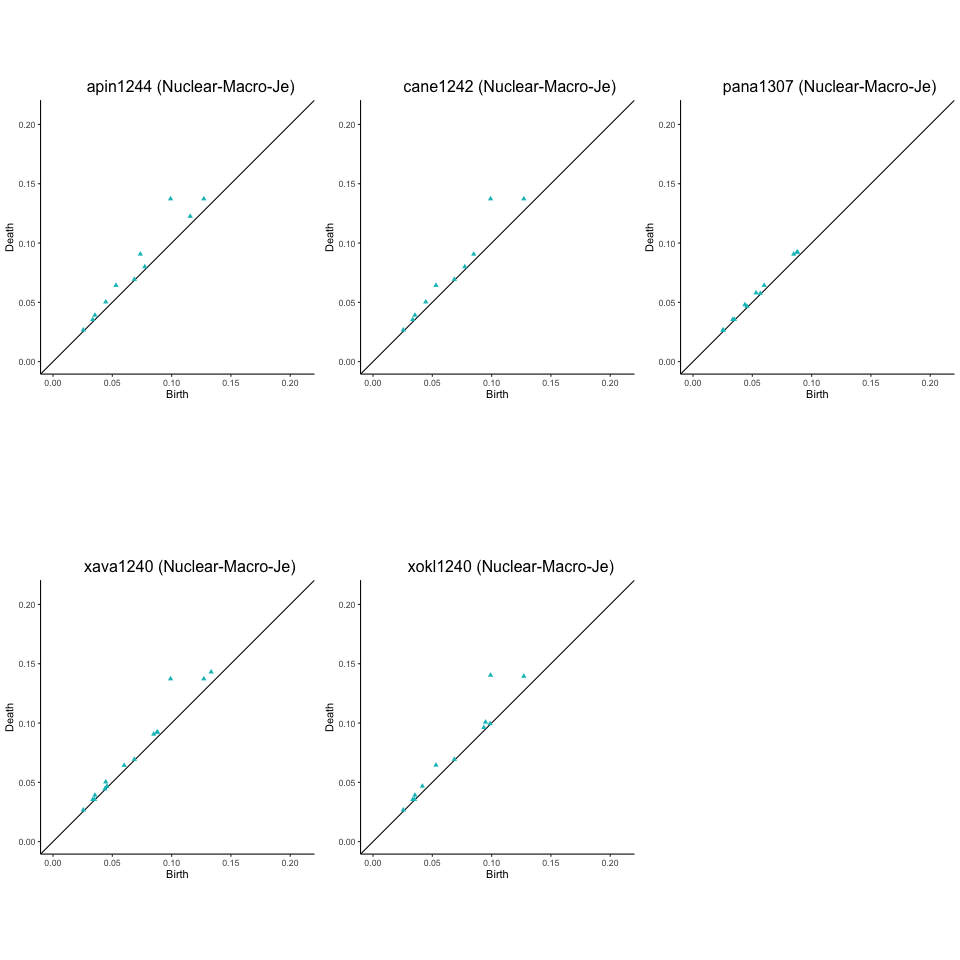}
    \caption{The persistence diagrams with respect to circular structures of J\^e-proper languages}
    \label{fig:pd_is_je}
\end{figure}

\begin{figure}
    \centering
    \includegraphics[width=0.4\textwidth]{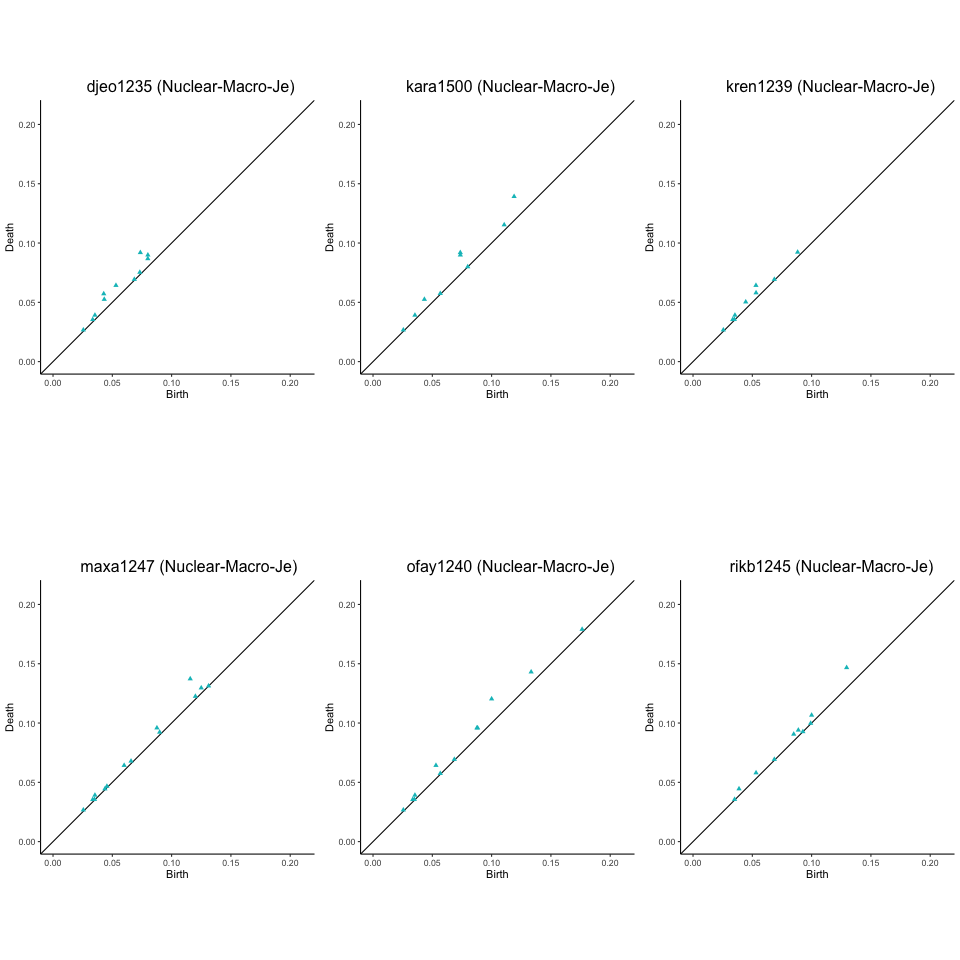}
    \caption{The persistence diagrams with respect to circular structures of non-J\^e-proper languages}
    \label{fig:pd_not_je}
\end{figure}

The persistence diagrams of J\^e-proper and the other NMJ languages are displayed in Figure \ref{fig:pd_is_je} and Figure \ref{fig:pd_not_je}, respectively.
We observe that there is a special circular structure in the J\^e-proper group when birth is around $0.10$ and death is around $0.15$, the only exception is the language pana1307 ( Panar\'a),
however there is no such a significant circular structure in the group of non-J\^e-proper languages.
As mentioned, Panar\'a (pana1307) forms an exception within the J\^e-proper group. 
The language has been described as having a deviant morphosyntactic profile compared to its sister languages 
It is at this point unclear what the sociohistorical explanation of this deviance is, but it is possibly related to a turbulent past in the last century or so, 
involving long-distance migrations and a severe population bottleneck as the result of exposure to diseases and ecological disruption of their habitat \cite{dourado_phd_panara},
\cite[page 11-12]{nikulin_proto_mj}.


We can detect which points contribute to this unique circular structure in each sub-point cloud using the R package ``TDA'' \cite{r_TDA},
the points are listed as below:
\begin{itemize}
\item apin1244: GB044:0, GB053:0, GB079:1, GB084:0, GB107:0, GB028:1, GB035:1, GB093:0, GB039:0, GB318:1, GB089:0, "GB136:1". 
\item cane1242: GB044:0, GB068:1, GB053:0, GB107:0, GB028:1, GB084:0, GB093:0, GB039:0, GB318:1, "GB136:1.
\item xava1240: GB044:0, GB042:0, GB053:0, GB039:0, GB079:1, GB084:0, GB107:0, GB093:0, GB318:1, GB193\_YX:1, GB089:0, GB136:1.
\item xokl1240: GB044:0, GB068:1, GB021:0, GB107:0, GB084:0, GB318:1, GB136:1.
\end{itemize}
If we take the intersection of all the points, 
we notice that the five points GB044:0, GB084:0 GB107:0, GB318:1 and GB136:1 appear in all four languages, while
in all the other languages there is not such a combination.
In  Panar\'a,
the values of these five features are: GB044:1, GB084:0, GB107:0, GB318:0 and GB136:1. 
The mini profile of these five feature values can be connected to the isolating tendencies of the J\^e-proper languages, and their dependence on syntactic rather than morphological strategies \cite[180]{rodrigues1999}, as they concern the overt marking of inflectional categories on nouns (plural) and verbs (future tense, negation), and the presence of fixed clausal word order. Panar\'a, having morphologized its number marking on nouns, is slightly less isolating than its sister languages, which use a prosodically free word to mark plural.

We can also measure the bottleneck distance and $2$-Wasserstein distance between different languages in NMJ.
This can be implemented by using 
the R package ``TDApplied'' \cite{r_TDApplied}. 
We then can apply the MDS method to visualize the corresponding distance matrix,
which is displayed in Figure \ref{fig:cloud}.
We notice that there is a clear gap between J\^e-proper group and non-J\^e-proper group.
\begin{center}
  \vspace{.1cm}
  \includegraphics[width=0.46\hsize]{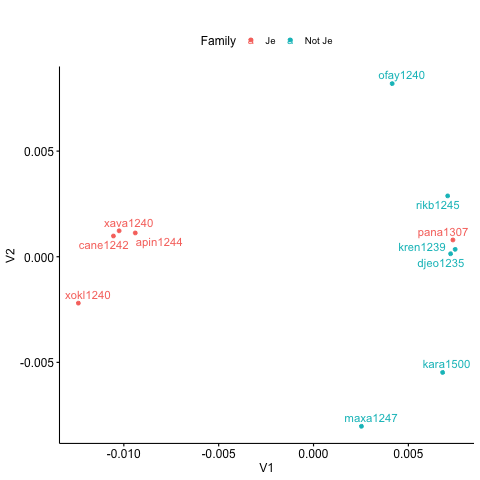}
  \includegraphics[width=0.46\hsize]{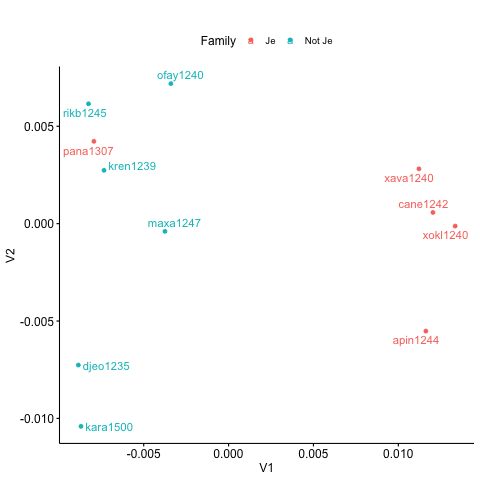}
   \captionof{figure}{Left: MDS plot w.r.t. bottleneck distance of Nuclear-Macro-J\^e. Right: MDS plot w.r.t. $2$-Wasserstein distance of Nuclear-Macro-J\^e.}
\label{fig:cloud}
  \nobreak
\end{center}

Moreover,
a hypothesis test for non-parametric statistical inference of persistence diagrams can be conducted.
We will use a permutation test method introduced in \cite{MR3975554} to test if the J\^e-proper group and the non-J\^e-proper groups are significantly distinguishable.

Our null hypothesis is that the sets of persistence diagrams associated with each group are indistinguishable.
The first step is to define a function that can quantify the difference between two groups (In terms of mathematical optimization or decision theory, this is called a loss function) based on the chosen measurement between two persistence diagrams.
Here we choose the bottleneck distance measurement and $2$-Wasserstein distance measurement separately.
Then we calculate the value of the loss function,
which is referred to as the observed value.
The next step is to randomly shuffle all the persistence diagrams multiple times and calculate the corresponding values of loss function.
Finally we evaluate the ratio of values that are less than or equal to the observed value, which accounts for the $p$-value of our permutation test.
There are many R packages like ``TDApplied'' and ``TDAstats'' which can implement the permutation test,
in this experiment we choose ``TDApplied'' and shuffle $100$ times.
in the case of $2$-Wasserstein distance, 
we get a $p$-value of $0.0297(=3/101)$,
and in the case of Bottleneck distance,
we get a $p$-value of $0.0099(=1/101)$.
We observe that in both cases our $p$-value is less than the threshold $0.05$,
hence our null hypothesis is rejected.
\begin{remark}
In our computation,
the $p$-value $0.0297$ associated with $2$-Wasserstein distance is a round value of $3/101$,
which means that among all the $100$ times of shuffles,
only $2$ times we get a value of loss function less than or equal to the observed value.
And the $p$-value $0.0099$ associated with bottleneck distance is a round value of $1/101$,
which means that among all the $100$ times of shuffles,
there is not any value of loss function that is less than or equal to the observed value.

In fact,
since in the family of NMJ there are only $11$ samples which are divided into two groups of size $5$ and $6$ separately,
we have only $462$ different permutations,
the calculation is relatively simple and we can get the exact $p$-value.

By computing the loss function over all the $462$ permutations,
we realize that in the measurement of $2$-Wasserstein distance,
$9$ out of $462$ values of loss function are less than or equal to the observed value,
hence the exact $p$-value is equal to $0.0195$,
and in the measurement of bottleneck distance,
$8$ out of $462$ values of loss function are less than or equal to the observed value,
the exact $p$-value is equal to $0.0173$.
\end{remark}

\subsection{Quechuan}\label{sec:quech}
The Quechuan language family stretches along the Andean mountain range from southern Colombia to northern Chile and Argentina, and contains about 40 languages. Most of its current spread was achieved only relatively recently, with the expansion of the Inca empire (13th-16th century CE), of which Quechua was the language of administration. The core of Quechuan, however, in central Peru, is of greater time depth, estimated to about 2000 years \cite[page 168]{adelaar2004}. The family is divided into two main branches: Quechua I (in central Peru) and Quechua II (corresponding to the later expansions north- and southward). The northern Quechuan languages in north Peru, Ecuador and Colombia have replaced previously existing languages, and have in the process undergone substrate influence \cite{Muysken2021}, making them somewhat deviant grammatically from the rest of the family.

Grambank contains $9$ Quechuan languages,
of which $8$ are remained after the data preprocessing procedure.
Among these $8$ languages,
the languages Calder\'on Quechua (cald1236), Chimborazo Quechua (chim1302) and Imbabura Quechua (imba1240) are in the north,
and the languages San Mart\'in Quechua (sanm1289), Ayacucho Quechua (ayac1239), Cuzco Quechua (cusc1236), Huallaga Quechua (hual1241), North Jun\'in Quechua (nort2980) are in the south. North Jun\'in Quechua and Huallaga Quechua are part of the Quechua I branch, whereas the others are part of the Quechua II branch. The northern (Ecuadorian) Quechuan languages form a sub-branch within Quechua II. 

\begin{figure}
    \centering
    \includegraphics[width=0.55\textwidth]{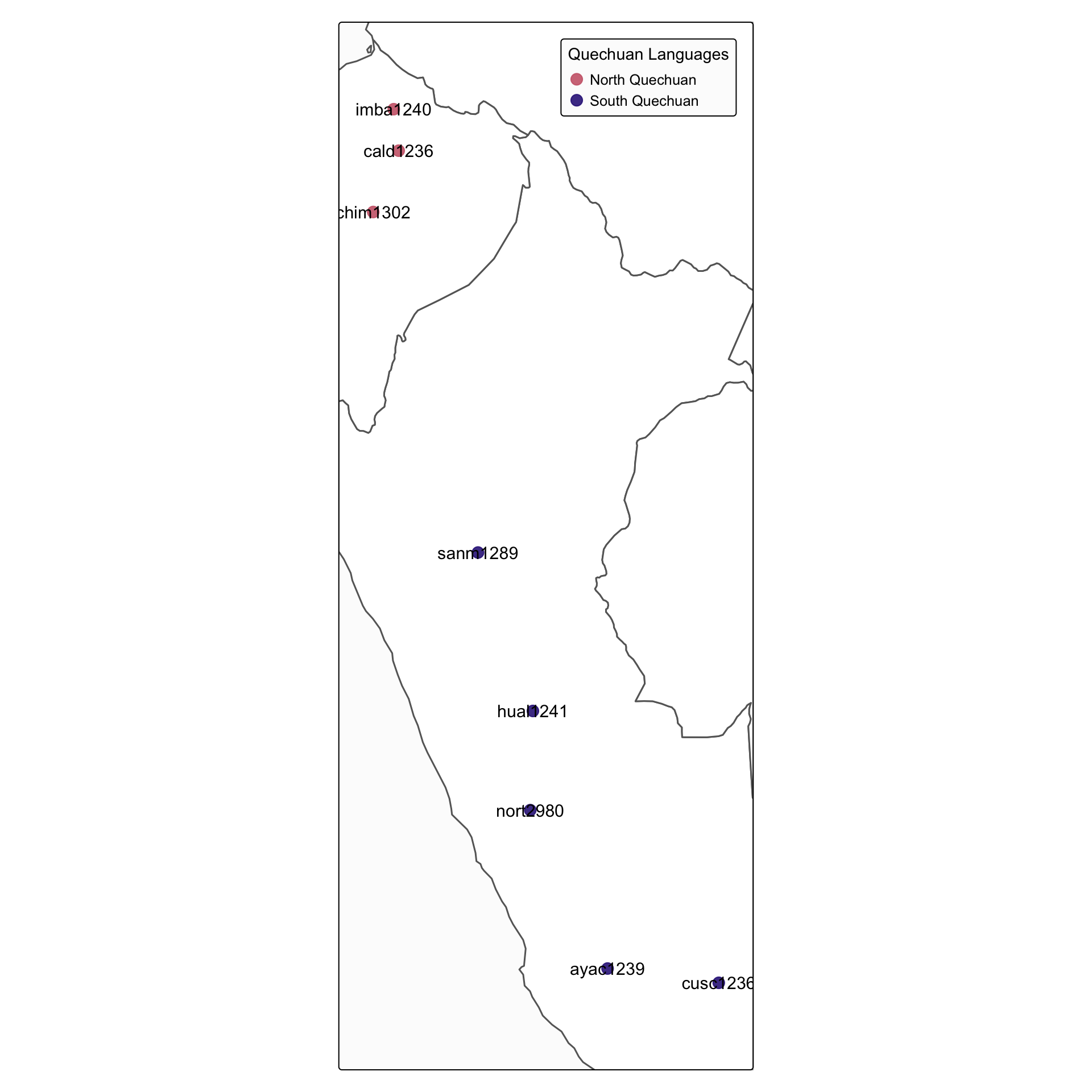}
    \caption{Quechuan Languages}\label{fig:quech_map}
\end{figure}
\begin{figure}
    \centering
    \includegraphics[width=0.4\textwidth]{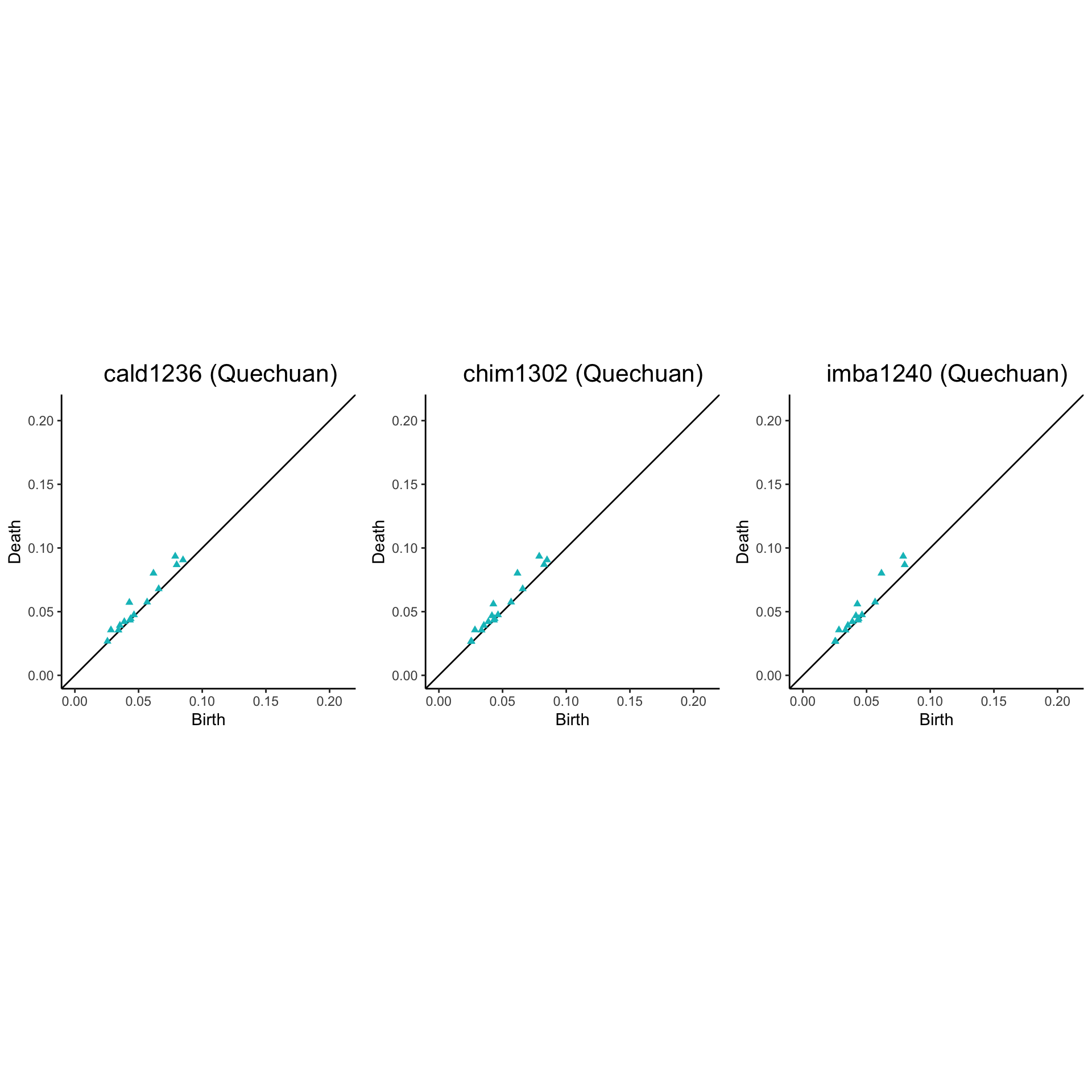}
    \caption{The persistence diagrams with respect to circular structures of Quechuan languages in the north}\label{fig:quech_pd_north}
\end{figure}
\begin{figure}
    \centering
    \includegraphics[width=0.4\textwidth]{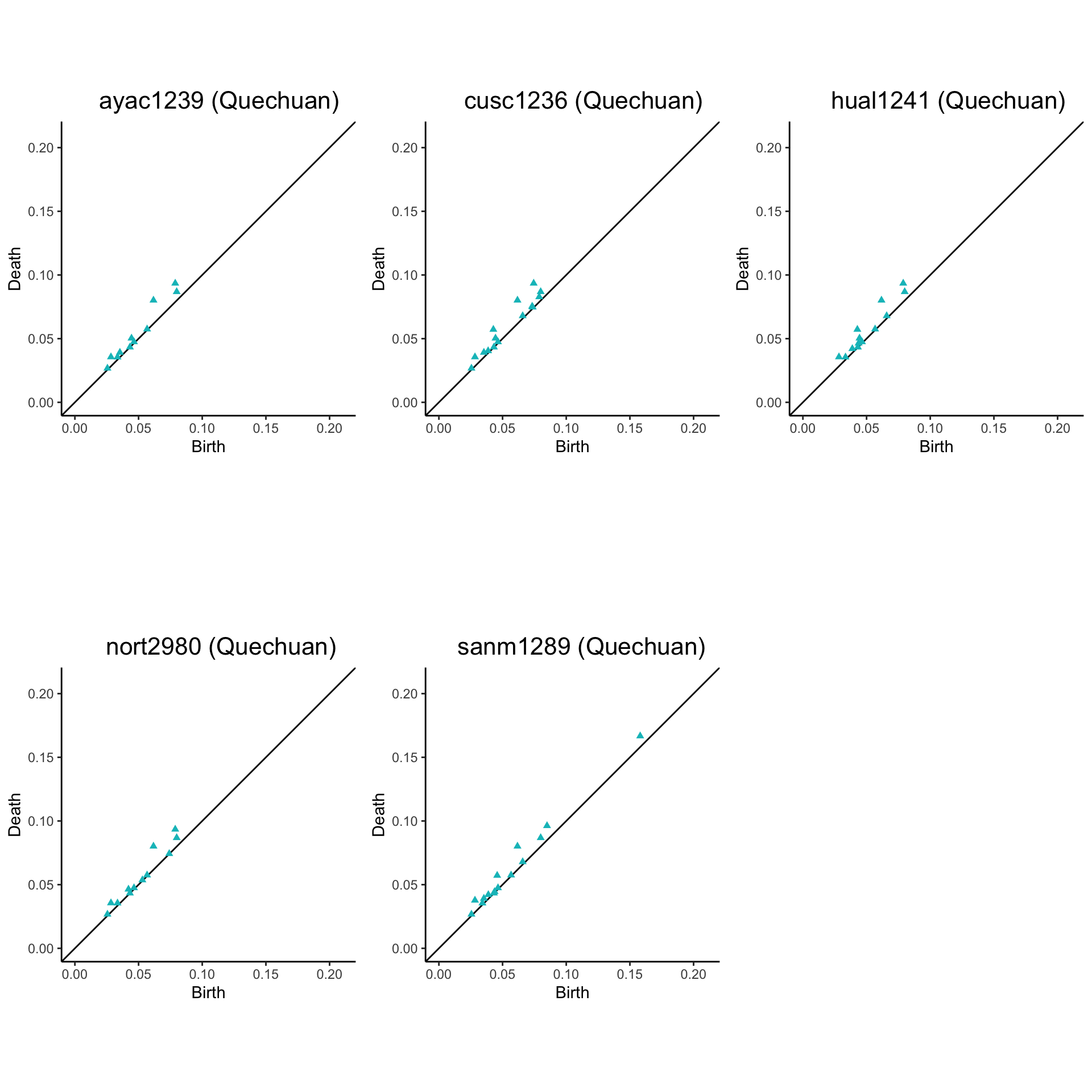}
    \caption{The persistence diagrams with respect to circular structures of Quechuan languages in the south}\label{fig:quech_pd_south}
\end{figure}
\begin{figure}
    \centering
    \includegraphics[width=0.4\textwidth]{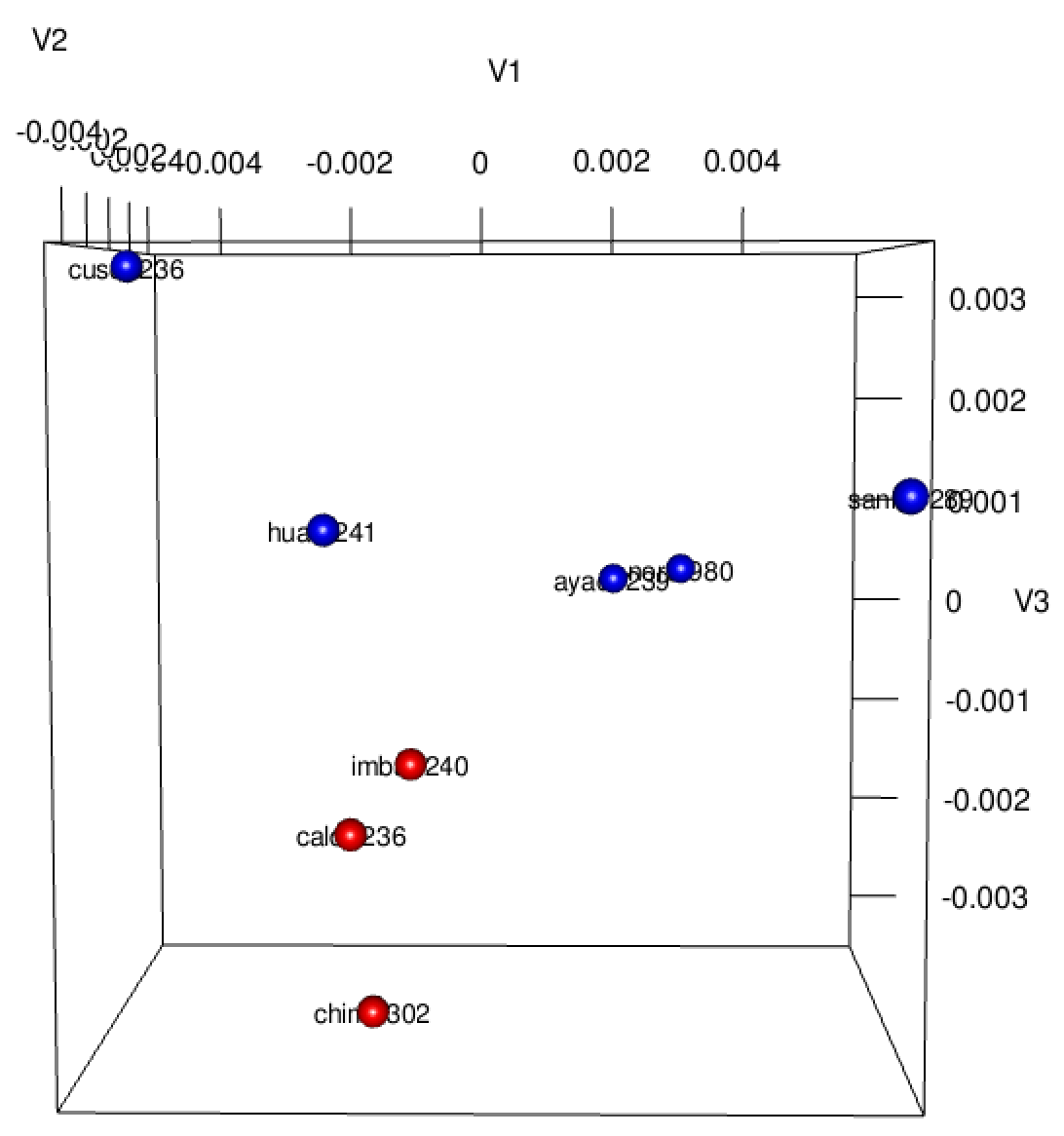}
    \caption{The $3$-dim MDS visualization of $2$-Wasserstein distance of Quechuan}\label{fig:quech_mds_3d}
\end{figure}

The persistence diagrams of Quechuan languages in north and south are shown in Figure \ref{fig:quech_pd_north} and Figure \ref{fig:quech_pd_south} respectively.
In the case of the Quechuan family
it is relatively difficult to detect the difference of the persistence diagrams just by visualization,
however,
we can measure the $p$-Wasserstein distance of each pair of languages.
We still choose $2$-Wasserstein measurement,
the Figure \ref{fig:quech_mds_3d} displays the $3$-dimensional MDS plot of Quechuan languages.
As it is shown,
the north Quechuan languages (red) and south Quechuan languages (blue) are separated into two different groups.
We then do a permutation test to make a decision as well.
Our null hypothesis is that the north Quechuan languages and south Quechuan languages are indistinguishable. 
In this example since the size of the two groups are $3$ and $5$ respectively,
there are only $56$ different permutations in total,
hence it is possible to calculate the loss function of each permutation.
Finally we obtain that the exact $p$-value is equal to $0.0179$.
If we set our threshold to be $0.05$,
our null hypothesis is then denied.
In fact,
the $p$-value $0.0179$ is equal to $\frac{1}{56}$,
which means that every non-trivial permutation gives a value of loss function larger than the observed value.

\begin{remark}
    It is worth to emphasize it again that the persistence diagrams in Figure \ref{fig:family_pd} are different from those in 
    Figure \ref{fig:pd_is_je},  Figure \ref{fig:pd_not_je}, Figure \ref{fig:quech_pd_north} and Figure \ref{fig:quech_pd_south},
    The reason is that in Figure \ref{fig:family_pd} we take the first two components of MCA projection,
    therefore all the point clouds are inside a $2$-dimensional Euclidean space,
    while in the rest cases we take the first four components of MCA projection,
    and the point clouds are inside a $4$-dimensional Euclidean space.
\end{remark}

\section{Discussion}
We have presented the visualization of languages via MCA method.
More than that, 
we showed how to apply the TDA framework to analyze the shapes of languages in the case of Nuclear-Macro-J\^e (NMJ) family and Quechuan family respectively.
In addition, we measured the bottleneck distance and $2$-Wasserstein distance of persistence diagrams.
We see that there is a significant difference between J\^e-proper and non-J\^e-proper via the $2$-dim MDS plot.
In the case of the Quechuan family,
we can see the difference between north Quechuan languages and south Quechuan languages via the $3$-dim MDS plot with respect to $2$-Wasserstein measurement.
Furthermore,
if we apply the permutation test,
by computing the $p$-value we see that in the case of the NMJ family,
J\^e-proper and non-J\^e-proper are significantly different from each other,
and in the case of the Quechuan family,
the north Quechuan languages and south Quechuan languages are significantly different as well.

The drawback of persistence diagram is that it contains only topological information.
In an extreme case,
suppose if the corresponding point clouds of two languages are congruent,
i.e.,
one point cloud can be transformed into the other point cloud by a rigid transformation (i.e., a composition of rotations, translations and reflections),
the two persistence diagrams would be exactly the same.
Nevertheless,
in the framework of topological data analysis
there are many more powerful tools like ``persistent homology transform'' \cite{MR3311455} that can deal with this issue.
Another drawback is that
the persistence diagram can detect only the topological structures like clustering structures and circular structures of a point cloud.
In contrast to that,
some recent works \cite{book_jost, pers_lap} indicate that the spectrum (or in other words, the eigenvalues) of the persistent Laplacian operators contain much more deeper geometrical information of a Vietoris-Rips filtration.
It would be an interesting topic to analyze what the spectrum of the persistent Laplacian operators can tell about a language.

\section*{Acknowledgements}
This work is part of the ERC Consolidator project South American Population History Revisited, 
funded by the European Research Council (ERC) under the European Union’s Horizon 2020 research and innovation programme (grant agreement No. 818854 - SAPPHIRE).
This support is gratefully acknowledged.
I would like to express my sincere gratitude to Rik van Gijn for the valuable discussion about South American languages and proofreading this work.
The main material of this work is based on a presentation I gave in the 2023 Grambank Workshop in Leipzig,
I feel thankful to Russell Gray and Hedvig Skirg\r{a}rd for their insightful suggestions.
In addition,
I also would like to express my thanks to Leonardo Arias Alvis and Sietze Norder for their inspiring feedback.


\appendix

\section{Multiple correspondence analysis}\label{appdx_mca}
In this work we use the multiple correspondence analysis (MCA) based on the Burt Matrix with adjustment.
We follow the notation of \cite{mjca}.
For a more complete introduction to MCA,
we refer the readers to \cite{mjca, MCA}.

Consider a table of $Q$ categorical features and $I$ samples,
we denote by $J_q$ the amount of categorical values the  $q$-th feature contains,
then in total we have $J=\sum_{q=1}^Q J_q$ different categorical values.
We then create an $I \times J$ matrix $Z$,
each row corresponds to a sample,
each column corresponds to a categorical value of a feature.
The entries of matrix $Z$ are either $0$ or $1$,
depends on the feature values of samples,
if a sample has a feature value, 
the corresponding entry is $1$,
otherwise,
it is $0$.
The matrix $Z$ is called the indicator matrix.
We then define the Burt matrix $B$ as $B=\{b_{ij}\}=Z^TZ$.

The MCA algorithm is implemented in the following steps:
\begin{enumerate}
\item Compute the correspondence matrix $P$:
$$
P = \{p_{ij}\} = {b_{ij}\Big/\sum_{i, j} b_{ij}}.
$$

\item Compute the row totals $r_i$:
$$r_i=\sum_{j}p_{ij}.$$

\item Compute the standardized residuals matrix $S=\{s_{ij}\}$ with 
$s_{ij}=(p_{ij}-r_ir_j)\big/\sqrt{r_ir_j}.$

\item Perform the singular value decomposition (SVD) on $S$:
$S=V\Lambda V^T,$
here $\Lambda=\diag\left(\lambda_1, \lambda_2, \cdots, \lambda_J\right)$ with $\lambda_1 \geq \lambda_2 \geq \cdots \geq \lambda_J$.

\item Compute the adjusted variance $\widetilde{\lambda}_j^2$:
\begin{equation*}
\widetilde{\lambda}_j^2
=
    \begin{cases}
      \left(\frac{Q}{Q-1}\right)^2 \left(\lambda_j-\frac{1}{Q}\right)^2 & \text{if $\lambda_j > \frac{1}{Q}$},\\
      0 & \text{if $\lambda_j \leq \frac{1}{Q}$}.
    \end{cases}       
\end{equation*}

\item Compute the principal coordinates of all samples
$F = D_r^{-\frac{1}{2}} V \widetilde\Lambda$,
here $D_r^{-\frac{1}{2}}$ and $\widetilde\Lambda$ are diagonal matrices defined by 
$D_r^{-\frac{1}{2}} = \diag(r_1^{-1/2}, r_2^{-1/2}, \cdots, r_J^{-1/2})$ and $\widetilde\Lambda = \diag\left(\widetilde{\lambda}_1, \widetilde{\lambda}_2, \cdots, \widetilde{\lambda}_J\right)$.

\item Compute the adjusted total variance:
$$\tau=\frac{Q}{Q-1}\left(\sum_{j=1}^J\lambda_j^2 - \frac{J-Q}{Q^2}\right).$$

\item Evaluate the percentage of variance for the $j$-th principal component: $\widetilde{\lambda}^2_j\Big/\tau$.
\end{enumerate}

\section{Background about topological data analysis}\label{appdx_tda}
Suppose $P$ is a point cloud of finitely many points inside a Euclidean space $\mathbb{R}^d$.
In this appendix we will give the main definitions and results necessary in this paper.
We follow mainly the notation of the book \cite{MR4381505}.
We refer to the readers \cite{MR2572029, MR4381505, MR2549932} for more details about the framework of TDA.

\subsection{Simplices and complexes}

\begin{definition}
We say a set of points $\{p_0, p_1, ..., p_k\}\subset \mathbb{R}^d$ is affinely independent if for any set of real numbers $\{t_0, t_1, ..., t_k\}\subset \mathbb{R}$,
$$
\sum_{i=0}^kt_i =0 \textrm{  and  }
\sum_{i=0}^kt_i p_i = 0
$$
implies 
$t_0=t_1=...=t_k=0$.
\end{definition}
The set $\{p_0, p_1, ..., p_k\}$ is affinely independent if and only if the $k$ vectors $\{p_1-p_0, p_2-p_1, ..., p_k - p_{k-1}\}$ are linearly independent.

\begin{definition}[Simplex]
If the set of points $P=\{p_0, p_1, ..., p_k\}$ is affinely independent,
the convex hull of $P$ is called a $k$-simplex.
\end{definition}
\begin{remark}
In more detail,
the convex hull of $P$ is the smallest convex set that contains $P$.
Follow this definition,
a $0$-simplex is a point,
a $1$-simplex is a line segment,
a $2$-simplex is a triangle,
and a $3$-simplex is a tetrahedron.
\end{remark}

A $k$-simplex $\sigma$ is said to have dimension $k$.
For a subset $P'\subset P$ that contains $k'$ points,
the convex hull of $P'$ is a $k'$-simplex,
we call it a $k'$-face of $\sigma$.

\begin{definition}[Simplicial complex]\label{def_simp_cmplx}
We call $K$ a simplicial complex if it is a finite collection of simplices and satisfies the following two restrictions:
\begin{itemize}
    \item Each face of each simplex is contained in $K$.
    \item Take any two simplices $\sigma_1$ and $\sigma_2$ in $K$,
    the intersection $\sigma_1 \cap \sigma_2$ is either a face of both $\sigma_1$ and $\sigma_2$ or empty.
\end{itemize}
\end{definition}

We define the dimension of a simplicial complex $K$ to be the maximum dimension of all simplices in $K$,
and denoted it as $\dim(K)$.
We say $K$ is a simplicial $k$-complex if $\dim(K)=k$.

\subsection{Chains and boundary operators}
Let $K$ be a simplicial $k$-complex with $n_p$ being the number of $p$-simplices, 
according to Definition \ref{def_simp_cmplx} it is easy to see that $k\geq p \geq 0$.

\begin{definition}[Chains]
A $p$-chain $\gamma$ in a simplicial complex $K$ over the field $\mathbf{F}_2$ is defined to be a formal sum of $p$-simplices,
that is,
$\gamma=\sum_{i=1}^{n_p}c_i \sigma_i$ with $c_i\in \mathbf{F}_2$ and each $\sigma_i$ being a $p$-simplex.
\end{definition}
The set of all $p$-chains form a vector space over the field $\mathbf{F}_2$,
we can sum two $p$-chains $\gamma = \sum c_i\sigma_i$ and $\gamma^\prime = \sum c_i^\prime\sigma_i$ up to obtain another $p$-chain $\gamma+\gamma^\prime = \sum (c_i + c_i^\prime)\sigma_i$,
and we can also multiply a $p$-chain $\gamma$ by a number $c^\prime \in \mathbf{F}_2$ to get a new chain $c^\prime\gamma=\sum_{i=1}^{n_p}c^\prime c_i\sigma_i$.
We denote the set of all $p$-chains as $C_p(K)$.
Since the simplicial complex $K$ is clear from the context we always drop $K$ and use the notation $C_p$ to denote the set of all $p$-th chains.

There is a group homomorphism from $C_p$ to $C_{p-1}$ which is referred to as a boundary operator.
the definition is given as follows.
\begin{definition}[Boundary operator]
We take $\sigma=\langle v_0, ..., v_p\rangle$ to be a $p$-simplex and let
$$\partial_p(\sigma) := 
\sum_{i=0}^p\langle v_0, ..., \hat{v}_i, ..., v_p\rangle,
$$
where $\hat{v}_i$ indicates that $v_i$ is removed.
Since $C_p$ is a vector space,
we can then extend $\partial_p$ to $C_p$ and obtain a group homomorphism $\partial_p: C_p\to C_{p-1}$ as
$$
\partial_p \gamma
 =
 \sum_{i=1}^{m_p}c_i \partial_p (\sigma_i)
$$
for a $p$-chain $\gamma=\sum c_i \sigma_i\in C_p$.
The map $\partial_p$ is called a boundary operator.
\end{definition}

The most important property of the boundary operator $\partial_p$ is that the composition of two boundary operators is zero.
\begin{theorem}\label{thm_apdx_1}
For any $p$-chain $\gamma$,
$\partial_{p-1}\cdot \partial_{p}(\gamma)=0$.
\end{theorem}

We can then obtain the following sequence of group homomorphisms:
$$
0 = 
C_{k+1}\xrightarrow{\partial_{k+1}}
C_k\xrightarrow{\partial_{k}}
\cdots
C_1\xrightarrow{\partial_{1}}
C_0\xrightarrow{\partial_{0}}
C_{-1}=0.
$$
We refer to such a sequence as a chain complex.

\begin{definition}[Boundary group]
We call the image of $\partial_{p+1}$ the $p$-th boundary group,
and denote it as $\mathsf{B}_p$.
That is,
$
\mathsf{B}_p
=
\big\{
\gamma\in C_{p} \,\big|\, \partial_{p+1}\Tilde{\gamma} = \gamma \textrm{ for some }\Tilde{\gamma}\in C_{p+1} 
\big\}
$
\end{definition}

\begin{definition}[Cycle group]
We call the kernel of $\partial_p$ the $p$-th cycle group, 
and denote it as $\mathsf{Z}_p$.
In more detail,
$\mathsf{Z}_p
=
\big\{\gamma\in C_p \,\big|\, \partial_{p}\gamma=0\big\}.
$
\end{definition}

Since the composition of two boundary operators $\partial_{p}$ and $\partial_{p-1}$ is zero by Theorem \ref{thm_apdx_1},
we can realize that $\mathsf{B}_p \subset \mathsf{Z}_p$,
and the quotient group plays an important role in algebraic topology.

\begin{definition}[Homology group]
The $p$-th homology group is defined as the quotient group $\mathsf{H}_p:=\mathsf{Z}_p/\mathsf{B}_p$.
\end{definition}
In our context,
the homology group $\mathsf{H}_p$ is a vector space over the field $\mathbf{F}_2$,
the dimension of $\mathsf{H}_p$ is called the $p$-th Betti number,
we denote it as $\beta_p$.

\subsection{Basics about persistent homology}

\begin{definition}[Vietoris-Rips complex]
We define the Vietoris-Rips complex of a point cloud $P$ at threshold $r$ to be the simplicial complex
$$
\Rips_r(P)
=
\big\{\sigma \subset P \big | \textrm{diam}(\sigma)\leq 2r\big\},
$$
here $\textrm{diam}(\sigma)$ denotes the diameter of a subset $\sigma$,
i.e.,
$\textrm{diam}(\sigma)=\max\limits_{p_i, p_j\in \sigma}||p_i-p_j||$.
\end{definition}

\begin{definition}\label{def_filt}
We define the Vietoris-Rips filtration of $P$ to be the nested collection of Vietoris-Rips complexes of $P$:
$$
\RRips(P)=\{\Rips_r(P)\}_{r\in[0, \infty)}.
$$
\end{definition}

The Definition \ref{def_filt} indicates that if $r_1 < r_2$,
the Vietoris-Rips complex $\Rips_{r_1}(P)$ is a sub-complex of $\Rips_{r_2}(P)$, 
which induces a group homomorphism $h_{p}^{r_1, r_2}$ in the level of homology groups.
In more detail,
\begin{equation}
\begin{split}
\Rips_{r_1}(P)&\subset \Rips_{r_2}(P)\\
&\Big\Downarrow\\
\mathsf{H}_p(\Rips_{r_1}(P)) &\xrightarrow{h_{p}^{r_1, r_2}} \mathsf{H}_p(\Rips_{r_2}(P)).
\end{split}
\end{equation}

\begin{definition}[Persistent homology]
The $p$-th persistent homology groups are defined as the images of the group homomorphisms $h_{p}^{r_1, r_2}$,
that is to say,
$\mathsf{H}_p^{r_1, r_2}:=\im\, h_{p}^{r_1, r_2}$.
\end{definition}

\begin{definition}[Persistent Betti number]
We define the dimensions $\dim \mathsf{H}_{p}^{r_1, r_2}$ of the vector spaces $\mathsf{H}_p^{r_1, r_2}$ as the $p$-th persistent Betti numbers, 
and denote it as $\beta_{p}^{r_1, r_2}$.
\end{definition}

\begin{definition}[Birth and death]
Let $\xi$ be a nontrivial $p-$th homology class in $\mathsf{H}_p(\Rips_a(P))$.
For a number $r_2\leq a$,
we say $\xi$ is born at $\Rips_{r_2}(P)$ if $\xi \in \mathsf{H}_p^{r_2, a}$ while $\xi \notin \mathsf{H}_{p}^{r_1, a}$ for any $r_1 < r_2$.
Similarly,
for a number $s_1 \geq a$,
we say $\xi$ dies at $\Rips_{s_1}(P)$ if $h_{p}^{a, s_1}(\xi)$ does not vanish while $h_p^{a, s_2}(\xi)=0$ for any $s_2 > s_1$.
\end{definition}

We can visualize the birth and death of homology classes via the so-called persistence diagram.
For a nontrivial $p-$th homology class $\xi\in \mathsf{H}_p(\Rips_{r}(P))$, if it is born at $b$ and dies at $d$, we denote it as a point $(b, d)$ in the plane $\mathbb{R}^2$,
as is shown in Figure \ref{img_pretzel_dg}.

Suppose we have two point clouds $P_1$ and $P_2$,
we hope to work out some method to measure the difference between $P_1$ and $P_2$ via the corresponding persistence diagrams.
To start with, 
we review the definition of matching between two persistence diagrams.

\begin{definition}[Matching]\label{dfn_match}
Let $P_1$ and $P_2$ be two point clouds and $\dgm_{p, 1}$, $\dgm_{p, 2}$ be the corresponding persistence diagrams with respect to the $p$-th persistent homology.
We denote by $A_{p, 1}, A_{p, 2}$ the non-diagonal points in the persistence diagrams $\dgm_{p, 1}$ and $\dgm_{p, 2}$ respectively.
For a point $a_1\in A_{p, 1}$,
let $\overline{a}_1$ denote the nearest point (in the sense of Euclidean distance) of $a_1$ on the diagonal.
Define $\overline{a}_2$ for every point $a_2\in A_{p, 2}$ similarly.
Let $\overline{A}_{p, 1}=\{\overline{a}_1\}_{a_1\in A_{p, 1}}$ and $\overline{A}_{p, 2}=\{\overline{a}_2\}_{a_2\in A_{p, 2}}$.
Let $\Tilde{A}_{p, 1} = A_{p, 1} \bigcup \overline{A}_{p, 2}$ and $\Tilde{A}_{p, 2} = A_{p, 2} \bigcup \overline{A}_{p, 1}$.
We call a bijective map $\varphi$ from $\Tilde{A}_{p, 1}$ to $\Tilde{A}_{p, 2}$ a matching.
\end{definition}

\begin{definition}[Wasserstein distance and bottleneck distance]\label{dfn_wass}
Let $\Phi$ be the set of all matchings as defined in Definition \ref{dfn_match}.
For any $q\in [1, +\infty]$,
the $q-$Wasserstein distance is defined as 
\begin{equation*}
\begin{aligned}
&d_q(\dgm_{p, 1}, \dgm_{p, 2})\\
=
&\inf_{\varphi\in \Phi}\left(\sum_{x\in \dgm_{p, 1}}(||x - \varphi(x)||_q)^q\right)^{1/q}.
\end{aligned}
\end{equation*}
Especially when $q=+\infty$ the distance $d_\infty(\cdot, \cdot)$ is referred to as the bottleneck distance.
\end{definition}

The significance of Wasserstein distance and bottleneck distance is that a ``slight modification'' of the point cloud results in only a ``slight modification'' of the corresponding persistence diagram.
We denote by $|P|$ the cardinality of the point cloud $P$.
In the following two theorems we assume $P_1$ and $P_2$ are two point clouds with $|P_1|=|P_2| = M < \infty$.

\begin{theorem}[Bottleneck stability theorem]
Let $P_1$ and $P_2$ be two point clouds,
if there is a number $\epsilon > 0$ and a bijection $\varphi: P_1\to P_2$ such that $||x - \varphi(x)|| \leq \epsilon$ for all $x\in P_1$.
Then 
$$
d_\infty(\dgm_{p, 1}, \dgm_{p, 2}) \leq \epsilon.
$$
\end{theorem}

\begin{definition}
Let $\Phi$ be the set of all bijections between point clouds $P_1$ and $P_2$.
We define the point cloud Wasserstein distance between $P_1$ and $P_2$ as
$$
W_p(P_1, P_2)=\inf_{\varphi\in \Phi}\left(\sum_{x\in P_1}||x-\varphi(x)||_p^p\right)^{1/p}.
$$
\end{definition}

In the case of Wasserstein distance we also have the following Wasserstein Stability theorem \cite{skraba2023wasserstein}:
\begin{theorem}[Wasserstein stability theorem \cite{skraba2023wasserstein}]
$$
d_q(\dgm_{p, 1}, \dgm_{p, 2}) \leq 2^{M/(q+1)}W_q(P_1, P_2).
$$
\end{theorem}

\subsection{Permutation test}
We briefly repeat the procedure of permutation test based on TDA.
The reader can check \cite{MR3975554} for a more detailed discussion.

Let $\mathbf{P}_1=\{P_{1, 1}, ..., P_{1, n_1}\}$ and $\mathbf{P}_2=\{P_{2, 1}, ..., P_{2, n_2}\}$ be two groups of point clouds.
Our purpose is to detect if the shapes of the point clouds $\mathbf{P}_1$ and the point clouds $\mathbf{P}_2$ are significantly distinguishable. 
We first compute the persistence diagrams with respect to the $p$-th persistent homology of each point cloud,
and denote the corresponding sets of persistence diagrams by 
$\mathbf{D}_1 = \{D_{1, 1}, ..., D_{1, n_1}\}$ and $\mathbf{D}_2 = \{D_{2, 1}, ..., D_{2, n_2}\}$.

We then need to pick up a Wasserstein distance $d_q(\cdot, \cdot)$ between two persistence diagrams.
Let $\avg(\mathbf{D}_1)$ be the average distance among all the persistence diagrams in the group $\mathbf{D}_1$,
i.e.,
$$
\avg(\mathbf{D}_1) 
=
\frac{1}{n_1(n_1-1)}\sum_{i, j=1}^{n_1}d_q(D_{1, i}, D_{1, j}),
$$
and $\avg(\mathbf{D}_2)$ is defined similarly for the set of persistence diagrams $\mathbf{D}_2$.

We define a loss function $F$ of $\mathbf{D}_1$ and $\mathbf{D}_2$ as
$$
F(\mathbf{D}_1, \mathbf{D}_2)
 = 
 \frac{1}{2}\left(\avg(\mathbf{D}_1) + \avg(\mathbf{D}_2)\right).
$$

We then randomly shuffle all the persistence diagrams in $\mathbf{D}_1$ and $\mathbf{D}_2$ together, 
and form two new groups $\widetilde{\mathbf{D}}_1$, $\widetilde{\mathbf{D}}_2$ of size $n_1$, $n_2$ respectively,
then compute the value $F(\widetilde{\mathbf{D}}_1, \widetilde{\mathbf{D}}_2)$ and check whether $F(\widetilde{\mathbf{D}}_1, \widetilde{\mathbf{D}}_2) \leq F(\mathbf{D}_1, \mathbf{D}_2)$ or not.

We repeat this shuffling procedure for $N$ times,
if among all the $N$ times of experiments,
$Z$ out of $N$ times it happens that $F(\widetilde{\mathbf{D}}_1, \widetilde{\mathbf{D}}_2) \leq F(\mathbf{D}_1, \mathbf{D}_2)$,
the $p$-value is then defined to be $\frac{Z+1}{N+1}$.

Intuitively speaking,
imagine if all the point clouds in $\mathbf{D}_1$ are identical and all the point clouds in $\mathbf{D}_2$ are identical as well while $\mathbf{D}_1$ are different from $\mathbf{D}_2$,
then $F(\mathbf{D}_1, \mathbf{D}_2)=0$ while $F(\widetilde{\mathbf{D}}_1, \widetilde{\mathbf{D}}_2)\neq 0$ for every non-trivial permutation.
Thus in this case the counting value $Z$ is equal to $0$,
and the corresponding $p$-value is then equal to $\frac{1}{N+1}$.




\bibliographystyle{plain} 
\bibliography{references}






\end{document}